\documentclass[letterpaper, 10 pt, conference]{ieeeconf}  % Comment this line out if you need a4paper

\IEEEoverridecommandlockouts                              % This command is only needed if 
\usepackage[utf8]{inputenc} % allow utf-8 input
\usepackage[T1]{fontenc}    % use 8-bit T1 fonts
\usepackage{hyperref}       % hyperlinks
\usepackage{url}            % simple URL typesetting
\usepackage{booktabs}       % professional-quality tables
\usepackage{amsfonts}       % blackboard math symbols
\usepackage{nicefrac}       % compact symbols for 1/2, etc.
\usepackage{microtype}      % microtypography
\usepackage[noadjust]{cite}
\usepackage[textwidth=2cm,colorinlistoftodos,disable]{todonotes}
\usepackage{textcase,microtype,marvosym}
\usepackage{amsmath,amssymb,graphicx,MnSymbol}
\usepackage{mathtools}
\usepackage{subfigure}
\usepackage{graphics} % for pdf, bitmapped graphics files
\usepackage{epsfig} % for postscript graphics files
\usepackage{nicefrac}
\usepackage{booktabs,colonequals}
\usepackage{comment}
\usepackage{tikz,pgfplots,pgfplotstable}
\pgfplotsset{compat=newest}
\pgfplotsset{plot coordinates/math parser=false}

\definecolor{lred}{RGB}{200,0,0} %light red?
\definecolor{dred}{RGB}{130,0,0} %dark red?

\definecolor{dblu}{RGB}{0,0,130} %dark blue
\definecolor{dgre}{RGB}{0,130,0} %dark green

\definecolor{dgra}{RGB}{50,50,50}
\definecolor{mgra}{RGB}{100,100,100}
\definecolor{lgra}{RGB}{220,220,220}
\definecolor{MPG}{RGB}{000,125,122}
\definecolor{ora}{RGB}{255,153,51}

\renewcommand{\vec}{\boldsymbol}

\newcommand{\vw}{\vec{w}}

\usetikzlibrary{arrows,shapes,plotmarks}

\tikzset{>=stealth'}
\tikzstyle{graphnode} =
   [circle,draw=black,minimum size=22pt,text centered,text
     width=22pt,inner sep=0pt]
\tikzstyle{var}   =[graphnode,fill=white]
\tikzstyle{obs}   =[graphnode,fill=black,text=white]
\tikzstyle{fac}   =[rectangle,draw=black,fill=black!25,minimum size=5pt]
\tikzstyle{facprior} =[rectangle,draw=black,fill=black,text=white,minimum size=5pt]
\tikzstyle{edge}  =[draw=white,double=black,thick,-]
\tikzstyle{prior} =[rectangle, draw=black, fill=black, minimum size=
5pt, inner sep=0pt]
\tikzstyle{dirprior} = [circle, draw=black, fill=black, minimum
size=5pt, inner sep=0pt]

\DeclareSymbolFont{stmry}{U}{stmry}{m}{n}
\DeclareMathSymbol\leftarrowtriangle\mathrel{stmry}{"5E}
\DeclareMathSymbol\rightarrowtriangle\mathrel{stmry}{"5F}

\usepackage{algorithm}
\usepackage{algorithmic}

\newcounter{ALC@tempcntr}% Temporary counter for storage
\newcommand{\gd}{\texttt{GD}}
\newcommand{\adam}{\texttt{Adam}}
\newcommand{\metagd}{\texttt{MetaGD}}

\newcommand{\metagdmemadam}{\texttt{MetaGDMemAdam}}
\newcommand{\ltolbgd}{\texttt{L2LBGDBGD}}
\newcommand{\loss}{\mathcal{L}_{\text{task}}}

\newcommand{\meta}{\hat{\eta}}

\renewcommand{\baselinestretch}{0.99}

\title{Online Learning of a Memory for Learning Rates}
\author{
  Franziska Meier$^{*, 1, 2}$, Daniel Kappler$^{*, 1}$ and Stefan Schaal$^{1, 3}$\\
  \thanks{$^{*}$ both authors contributed equally}%
    \thanks{$^{1}$Autonomous Motion Department, MPI-IS, T\"ubingen,
    Germany.}%
  \thanks{$^{2}$RSE-Lab, University of Washington, Seattle, USA.}%
  \thanks{$^{3}$CLMC-Lab, USC, Los Angeles, USA.}%
  \thanks{Acknowledgements: This research was supported in part by National Science Foundation grants IIS-1205249, IIS-1017134, EECS-0926052, the Office of Naval Research, the Okawa Foundation, and the Max-Planck-Society. 
  }
}

\begin{document}

\maketitle
% THIS HAS TO BE REMOVED IT IS JUST TO SEE THE FUCKING PAGE NUMBERS FOR ONCE.
%thispagestyle{plain}
%\pagestyle{plain}

%===============================================================================

\begin{abstract}
  The promise of learning to learn for robotics rests on the hope that
  by extracting some information about the learning process itself we
  can speed up subsequent similar learning tasks. Here, we introduce a
  computationally efficient online meta-learning algorithm that builds
  and optimizes a memory model of the optimal learning rate landscape
  from previously observed gradient behaviors. While performing task
  specific optimization, this \emph{memory of learning rates} predicts
  how to scale currently observed gradients. After applying the
  gradient scaling our meta-learner updates its internal memory based
  on the observed effect its prediction had. Our meta-learner can be
  combined with any gradient-based optimizer, learns on the fly and
  can be transferred to new optimization tasks. In our evaluations we
  show that our meta-learning algorithm speeds up learning of MNIST
  classification and a variety of learning control tasks, either in
  batch or online learning settings.
\end{abstract}

%\keywords{Meta-Learning, motor skill learning} 

%===============================================================================

\section{Introduction}
The remarkable ability of humans to quickly learn new skills stems
from a hierarchical learning process. Instead of learning each new
skill from scratch, a higher-level -- more abstract -- meta-learner
acquires information about the learning process itself which is used
to guide and speed up the learning of new skills. For instance, it has
recently been shown that humans learn how much to correct for observed
motor skill errors, and reuse such a \emph{error sensitivity memory}
in subsequent skill adaptation tasks \cite{Herzfeld:2014es}. In a
sense, we are learning how to learn.

Most robotic learning tasks would benefit from being guided by such a
meta-learner, especially when we consider the incremental learning of
several skills. For example when learning to detect and recognize
certain objects, it should become easier to learn how to recognize new
object classes over time. The same is true for learning control tasks
such as learning task-specific models of a robot's dynamics
\cite{kappler_iros_2017}. Even single task learning settings can
benefit from such a meta-learning process. For instance, consider
robotic reinforcement learning tasks for which data-efficiency has
been a key challenge~\cite{deisenroth2011pilco, gu2017, Kamthe2017},
because acquiring new observations on a real system can be extremely
costly. Meta-learning processes can help to maximize the effect of
each acquired rollout.

In the machine learning community, this concept of learning-to-learn
has been explored in a variety of contexts
\cite{Hochreiter01learningto, schmidhuber1987evolutionary,
  schweighofer2003meta, vilalta2002perspective} and recently received
renewed attention \cite{andrychowicz2016learning, li2016learning,
  daniel2016learning, hansen2016using, fu2016deep}.
\begin{figure}
  \centering
  \includegraphics[width=0.8\columnwidth]{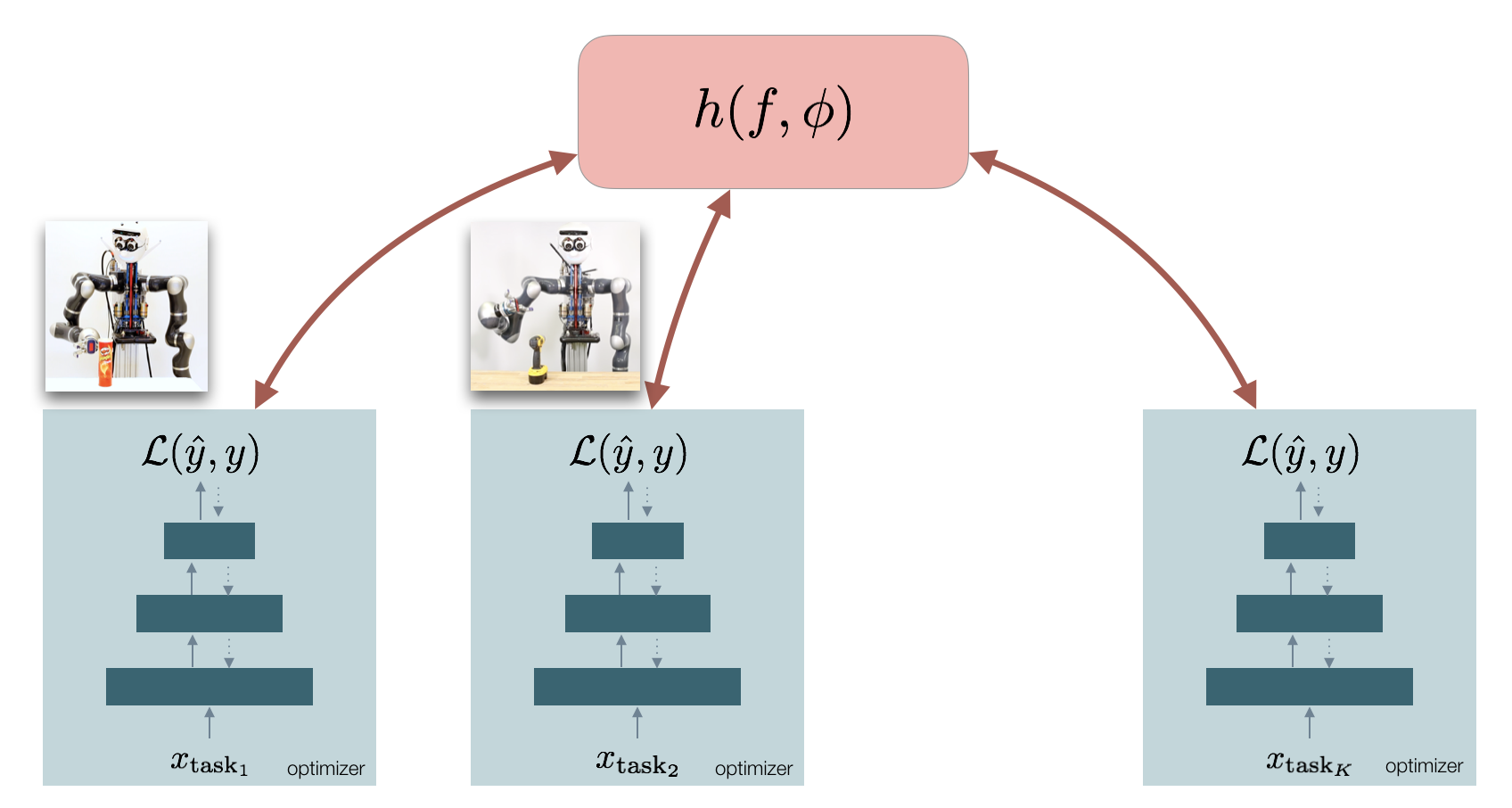}
  \caption{\small{Sequential learning of multiple task variants
      (lifting pringles, lifting drill) with meta-learning. In this
      work, the meta-learning process learns a \emph{memory of
        learning rates} $h$, which can be transfered between different
      learning problems. Transferring $h$ leads to faster learning of
      new task variations.}} \label{fig:overview}
\end{figure}
Recent work on learning how to optimize
\cite{andrychowicz2016learning, li2016learning, daniel2016learning,
  hansen2016using, fu2016deep} employs a two-phase approach: First a
meta-learner is optimized to perform well on a priori chosen tasks.
Then, once this meta-learner has been trained, it is utilized in
similar optimization tasks. In this work, we propose an online
meta-learning algorithm, that learns to predict learning rates given
currently observed gradients, while optimizing task-specific problems.
We show how we can train such a \emph{memory of learning rates} online
and in a computationally efficient manner. Our proposed approach has
several advantages:
Because the meta-learning is performed online, each observed data
point is utilized for both the task learning as well as the
meta-learning. Not only does this mean that we utilize observed data
more effectively, but also that the effect is immediate. Furthermore,
%performing the meta-learning online alleviates the need for having to
online meta-learning alleviates the need for having to
collect data on which to perform the learning-to-learn optimization
process. Thus, our meta-learner can improve when necessary, while
recent work is constrained to perform well on the task-distributions
it was trained on. Finally, the resulting memory of learning rates can
be transferred to similar optimization problems, to guide the learning
of the new task.

Here, we evaluate our approach on two supervised learning tasks:
sequential binary classification tasks on the MNIST \cite{MNIST} data
set, and incremental learning of a robot's inverse dynamics models.
Our experiments show that when combining an optimizer with our
meta-learner, we generally increase convergence speed, indicating that
we utilize observed data points more effectively to reduce errors in
the learning task. Furthermore, we show that when transferring our
meta-learner's internal state to new learning tasks, learning progress
is faster.

In the following we start out by presenting background and related
work in Section~\ref{sec:background}. We then introduce our
meta-learning approach in Section~\ref{sec:main}. Finally we both
illustrate and extensively evaluate our approach in
Section~\ref{sec:experiments} before concluding in
Section~\ref{sec:conclusion}.

\section{Background}\label{sec:background}
As mentioned above, most learning control problems use computationally
efficient variants of gradient descent
%In our work we specifically look at optimization problems $f(w)$ that
%are typically tackled via (stochastic) gradient descent
%
\begin{align}
	w^{t+1} = w^{t} - h(\nabla_w \loss(f(w)))
\end{align}
where $\loss$ typically corresponds to some loss function, $f$ the model with parameters $w$ to be optimized, and $h$ transforms the observed gradient
according to some rule. Note, in the remainder of this paper we sometimes suppress the dependency of $\loss$ on $f$, such that $\loss(f(w)) = \loss(w)$.

In this setting meta-learning can happen at several levels. For instance, recent work \cite{finn2017model} proposes a meta-learner that biases the learning process towards feature representations that supports few-shot learning of new tasks. On the other hand, recent learning to learn approaches \cite{andrychowicz2016learning, li2016learning, daniel2016learning, hansen2016using, fu2016deep} have been focused on learning optimizers that can be re-used for similar optimization tasks. The focus of these approaches however is on mitigating the issue of hyperparameter tuning instead of learning representations that can be transferred between (robotic) learning tasks.

Here, we present a novel meta-learning algorithm that is aligned with this second type of meta-learning, but is focused on maintaining an internal memory that can guide the learning of new tasks. In the following we review related work concerned with adaptive and learned optimizers.

\subsection{Adaptive First-Order Methods}
In the machine learning community adaptive optimizers such as \emph{Adam}
\cite{kingma2014adam}, \emph{RMSprop} \cite{2012Tieleman},
\emph{AdaGrad} \cite{duchi2011adaptive}, AdaDelta
\cite{2012arXiv1212.5701Z} have been developed. In essence, these
optimizers extend the gradient transform mapping $h$ to be a function
of sufficient statistics such as the mean and variance of the observed
gradients
\begin{align}
	w^{t+1} = w^{t} - h(\nabla_w \loss(w^t), \mathbb{E}[\nabla \loss(w)], \mathbb{E}[ (\nabla \loss(w))^2])
\end{align}
The specific gradient transform $h$ then depends on the algorithm, as
has been nicely summarized in \cite{daniel2016learning}, Table 1.
These optimizers are of linear space and time complexity with respect
to the model parameters, and thus are suitable for highly
parameterized models such as deep neural networks. 
Yet, choosing the
initial learning rate for each of these methods, or which base learner
to choose remains a complex manual tuning task
\cite{2016arXiv160803983L, 2017arXiv170602677G}. More importantly though, such adaptive methods are not meant to extract information about the learning process itself. Thus, while they have proven successful at increasing convergence speed, they are not designed to transfer knowledge to new optimization tasks.

\subsection{Learning to Learn}
More recently, the idea of meta-learning -- learning to learn --
has re-gained momentum \cite{andrychowicz2016learning, li2016learning,
  daniel2016learning, hansen2016using, fu2016deep}. These approaches
parametrize $h$ to be a function of parameters $\theta$, which
determine how observed gradients are to be transformed
\begin{align}
	w^{t+1} = w^{t} - h(\nabla_w \loss(w^t); \theta).
\end{align}
The goal then is to learn $\theta$ to create a well-performing
optimizer. In the most simple case, the parameter $\theta$ simply equals the
learning rate $\eta$, which can be adapted online, as has been shown
in \cite{2017arXiv170304782G}. Specifically, the authors propose to
compute the gradient $\nabla_\eta h$ 
online and then perform gradient
descent on the learning rate itself. While this approach is simple and
general as well as computationally and memory efficient, it does not
retain a state -- everything is forgotten. Thus subsequent
optimization tasks start from scratch.

When going from adaptive optimizers to learned optimizers we hope to
have learned something that we can reuse later on; that we performed
meta-learning on some level. Recent work such as
\cite{daniel2016learning, andrychowicz2016learning, li2016learning}
addresses this to some extent. An optimizer is trained to perform well on
some pre-defined set of optimization tasks and is then used to optimize similar
learning problems. While some approaches learn to transform the
gradient \cite{andrychowicz2016learning, li2016learning}, others
assume $h$ to be a scaling of the gradient and learn to predict the
learning rates \cite{daniel2016learning, hansen2016using, fu2016deep}.

Our work, falls into this second category of trying to learn a
coordinate-wise scaling of the gradient. To the best of our knowledge,
recent work either performs learning on the step size online, but
retains no memory \cite{2017arXiv170304782G}, or the learning of the
optimizer is performed once at the beginning and never updated again.
In the latter setting, two learning phases exist: In phase one $h$ is
trained, either via reinforcement learning \cite{daniel2016learning,
  li2016learning, hansen2016using, fu2016deep} or in a supervised
manner \cite{andrychowicz2016learning}; In phase 2, the trained
gradient transform $h$ is used to optimize similar learning
problems. 

These recent learning to learn approaches are mostly focused on mitigating the issue of hyper-parameter tuning by learning an optimizer. The goal of our approach is to learn a representation of optimal gradient transforms that can be transferred to new learning tasks. Furthermore, as opposed to previous work, we learn this representation in an online fashion while using this representation to optimize task-specific optimization problems.
\section{Learning a Memory for Learning Rates}
\label{sec:main}
In this work we investigate how we can learn a \emph{memory of
  learning rates} online, in a computationally efficient manner.
Ideally, this \emph{memory} can be transferred to subsequent similar
learning tasks, and speeds up convergence of that optimization
problem. Furthermore, this memory should be continuously updated to be
able to adapt and compress new learning rate landscapes as well.

We envision our meta-learning algorithm to be used as follows: While
our systems attempts to learn a model $f(w)$ for a specific task, such
as a task-specific inverse dynamics model, by minimizing the loss
$\loss(f(w))$ it also aims at building a model $h$ that can predict
how much to correct for observed errors. Thus for each optimization
step, we perform two updates: a gradient descent step on the
task-specific model parameters, and an update our meta-learner's
memory $h$. An overview in form of a pseudo algorithm is given in
\ref{alg:main}.

In order to develop such a meta learning approach several challenges
need to be met. One of the core challenges is the choice of training
signal to learn such a memory on. We take inspiration from
\cite{Riedmiller:1993jl} and start out by showing how to derive a
simple learning algorithm to train a learning rate memory $h$ for
one-dimensional optimization problems. Then, we discuss a
representation for $h$ that supports computationally efficient,
incremental learning. Finally we show how to generalize our meta learning approach to optimization problems involving complex models such as neural networks.

\subsection{Training Signal for the Learning Rate Memory}
\label{sec:mem_training}
Let us assume that our main objective is a learning task that requires
us to minimize the objective $\loss(w)$ with respect to parameter $w$.
While optimizing parameter $w$ we also aim to learn a function $h$
that given gradient information of the main objective $\loss(w)$, can
predict a learning rate $\hat{\eta}$
\begin{align}
  h(z; \theta) = \meta(z; \theta) z \label{eq:meta_def}
\end{align}
where $\theta$ are the parameters of the \emph{learning rate memory}
$h$. Ideally, we would like to optimize the parameters $\theta$ with
respect to the loss $\mathcal{L}_\text{lr}$
\begin{align}
  \mathcal{L}_\text{lr}(\eta, \meta(z; \theta)) = \frac{1}{2}(\eta - \meta(z; \theta))^2
\end{align}
where $\eta$ is the true optimal learning rate, and $\meta(z; \theta)$ the
predicted learning rate for input $z=\nabla_{w} \loss(w)$. To optimize
parameters $\theta$ via gradient descent, we would need access to the
true learning rate $\eta$
\begin{align}
  \frac{\partial \mathcal{L}_\text{lr}}{\partial \theta} = -(\eta - \meta(z; \theta)) \frac{\partial \meta(z; \theta)}{\partial \theta} 
\end{align}
which is unknown to us. However, by comparing the gradient of the
current time step with the gradient of the previous time step, we can
determine whether we over or underestimated $\eta$. If the gradient
has flipped between two consecutive optimization steps, the learning
rate was too large, meaning $\eta - \meta(z; \theta)>0$. On the other
hand, if the signs of the gradients are the same, then we can most
likely increase the learning rate.

With this knowledge at time step $t$, the memory parameter updates can be approximated as
\begin{align}
  \theta^{t+1} &= \theta^{t} - \xi \; (-(\eta^t - \meta(z^t; \theta^{t}))) \frac{\partial \meta(z^t; \theta)}{\partial \theta^t} \\
  &\approx \theta^{t} + \xi \; \text{sign}\left(z^{t+1} z^{t}\right) \frac{\partial \meta(z^t; \theta^t)}{\partial \theta}
\end{align}
where $\xi$ is a step size parameter for the gradient descent on
$\theta$. The exact form of this gradient update depends on what
parametric form $\meta(z; \theta)$ takes.

\setlength{\textfloatsep}{4pt}
\begin{algorithm}[t]
\caption{Online Meta Learning}\label{alg:main}
\begin{algorithmic}[1]
\REQUIRE initial task model $f(w^0)$, initial meta-memory $h_{\theta^{0}}$
\STATE $z^0 = 0$
\FOR{ $t \in {1,...,T}$}
\STATE $z^t = \nabla_w \loss(f(w^{t-1}))$
\STATE $h_{\theta^{t}}$ = MemoryUpdate($z^t, z^{t-1}, h_{\theta^{t-1}}$) \COMMENT{eq.~\ref{eq:mem_update}}
\STATE $w^{t} = w^{t-1} - h_{\theta^{t}}(z^t)$ \COMMENT{eq.~\ref{eq:meta_def} and eq.~\ref{eq:eta_hat}}
\ENDFOR
\RETURN $f(w^{T}), h_{\theta^{T}}$
\end{algorithmic}
\vspace{-0.1cm}
\end{algorithm}
\setlength{\textfloatsep}{6pt}
\subsection{Learning Rate Memory Representation}
\label{sec:mem_representation}
As mentioned previously, we aim at developing an algorithm that can
continuously update the learning rate memory $h$. Designing the
function approximator $h$, such that forgetting of previously learned
parameters is minimized, is one of the challenges of this approach.
Here we choose to use locally weighted regression
\cite{schaal1998constructive}, which is known for computational
efficiency and which has the capability to increase model complexity
when necessary. With locally weighted regression we decompose the
memory $h$ into $M$ local models $h_m$, each parametrized by their own parameters $\theta_m$.

Furthermore, in locally weighted regression, the loss function
$\mathcal{L}_{lr}$ also decomposes into $M$ separately weighted
losses, each dependent on only their respective parameters $\theta_m$:
\begin{align}
\mathcal{L}_{lr}(\theta) = \sum_{m=1}^M \mathcal{L}_m(\theta) = \sum_{m=1}^M \psi_m(z) (\eta - \meta_m(z, \theta_m))^2
\end{align}
where $\psi_m(z)$ is the weighting function that defines the active
neighborhood for each local model. A standard selection of this weight
function is the squared exponential kernel
$\psi_m(z) = \exp{(-0.5 \frac{(z - c_m)^2}{\lambda_m^2}})$.

Using this memory representation leads to following update rule per
local model
\begin{align}
  \theta^{t+1}_m &= \theta^{t}_m + \xi \; \text{sign}\left(z^{t+1} z^{t}\right) \psi_m(z^t) \frac{\partial \meta_m(z^t; \theta_m^t)}{\partial \theta_m^t}. \label{eq:mem_update}
\end{align}
Note, how only local models that are sufficiently activated require
updating. Finally, at prediction time the predicted learning rate is a weighted average over all local models' predictions:
\begin{align}
\hat{\eta} = \frac{\sum_{m=1}^M \psi_m(z) \meta_m(z, \theta_m)}{ \sum_{m=1}^M \psi_m(z) } \label{eq:eta_hat}
\end{align}
We now have an algorithm that can train a model $\meta$ to predict a
learning rate for one-dimensional optimization problems. An
illustration of what kind of learning rate landscapes can be trained
with this algorithm is given in the experimental Section in
Figure~\ref{fig:GD_vs_MetaGD_rb}.

As a final step, we show how this approach can be generalized to
models with multiple high-dimensional parameter groups, as commonly
encountered in deep learning models.
\subsection{Multi-Dimensional Learning Problems}
Above we have assumed $z = \nabla_w \loss(w)$ to be one-dimensional. A key
question is how to generalize to optimization tasks that not only have
high-dimensional gradients $\vec{z} = \nabla_{\vw} f(\vw)$, but also
multiple layers of parameters $\loss(\vw^1, \dots, \vw^K)$, where $\vw^k$
stands for the $k^{th}$ parameter group, as is typical for deep neural
networks for instance.

The straightforward extension would be to compute $\text{sign}(z^{t+1} z^t)$ as the inner product $\text{sign}({\vec{z}^{t+1}}^T \vec{z}^t)$ of two consecutive gradients, and place local models in the $D_k$-dimensional gradient space. However, this has the following consequences: We would only learn to
predict one learning rate per time step and we loose a lot of
information through the inner product of the two gradients. Furthermore, the number of local models needed to cover the gradient space evenly would grow exponentially with the gradient dimension $D_k$ and would significantly increase memory requirements while decreasing computational efficiency.

On the other end of the spectrum we can choose
$z = \nabla_{\vw_i} \loss(\vw)$ 
to be the partial derivative with respect to
the $i^{th}$ coordinate, and create a learning rate memory per
coordinate of the parameter vector $\vw$. Assuming that
$\vw \in \mathbb{R}^D$, this would require $D$ memory models, each
with $M$ local models. This choice, would create a learning rate
prediction per gradient coordinate, and thus offers maximum
flexibility. However, it also has high memory (storage)
requirements.

Here we choose a different route. First, we identify natural
parameter groups, such as parameters $\vw^k$ of each hidden
layer in deep neural networks. For each of these $K$ parameter groups,
a memory $h^k$ is created. Then, for the $k^{th}$ parameter group, we
pool the updates of all coordinate-wise updates, by computing the
average update per local model, across all coordinates
\begin{align}
a_{m,d}^{t,k} &= \text{sign}\left([z]_{d,k}^{t+1} [z]_{d,k}^{t}\right) \psi^k_m([z]_{d,k}^t) \\
  \theta^{t+1,k}_m &= \theta^{t,k}_m + \xi \; \frac{1}{D_k} \sum_{d}^{D_k}\big( a_{m,d}^{t,k} \; \frac{\partial \meta^k_m([z]_{d,k}^t; \theta^{t,k}_m)}{\partial \theta^k_m} \big).
\end{align}
where $[z]_{d,k}$ means we take the $d^{th}$ coordinate of
$z=\nabla_{w^k} \loss(\vw^{k})$, and $D_k$ denotes the dimensionality of the
$k^{th}$ parameter group $\vw^k$. At prediction time, we similarly make
predictions for each parameter group, per gradient coordinate $[z]_{d,k}$
to obtain a learning rate per parameter dimension.

Intuitively, this choice means that parameters within the same
parameter group share the same learning rate memory, expecting that
they would benefit from the same scaling behavior. With this
representation our meta-learner then requires extra memory resources
in the order of $O(KM)$. 
Furthermore, the computational complexity of the memory update
as well as learning rate prediction is in the order of $O(DM)$.
\subsection{Implementation Details}
Finally, to implement this approach, a few design choices have to be
made: Each memory is pre-allocated with a fixed number $M$ of local
models. Since, the localization happens in the space of
one-dimensional partial gradients - we linearly space the local models centers within the range of the minimum and maximum gradient value allowed\footnote{this corresponds to the choice of gradient clipping as is common in Tensorflow implementations}. The size of each local model is determined by parameters $\lambda_m$, which are chosen to create a reasonable amount of overlap between neighboring local models.
Thus the more local models we allow, the smaller they become.

Furthermore, we choose local constant models, such that
$\meta(z, \theta_m) = \theta_m$ in our experiments. Intuitively, this
means that each $\theta_m$ corresponds to a learning rate value,
localized in gradient space. We further use $z^{t+1} z^t$ and clip
the resulting (absolute) value at 1 instead
of the $\text{sign}(z^{t+1} z^t)$. 
We have empirically found that this significantly improves the 
convergence since memory updates are less pronounced in regions 
with small gradients.
Finally, at the beginning of a learning
problem, when no previous memory of learning rates exist, we
initialize the memories to predict an initial learning rate
$\eta_\text{init}$, thus at the very beginning all
$\theta_m = \eta_\text{init}$.

At runtime, when performing a task-specific learning problem, we then
immediately start optimizing all $h^k$ as well. At each time step $t$,
we update both, the parameters of the task-specific problem $\vw$ and
the learning rate memory parameters $\vec{\theta}$.

On a final note, we want to point out that our presented approach here
can be used on top of any base-learner. While not explicitly noted, it
is easily possible to first apply some fixed-rule based learner, such
as Adam \cite{kingma2014adam} to transform the gradient, and then
apply our learned memory evaluation on that transformed gradient. In
fact, in our experimental evaluation we include results for both basic gradient descent with learning rate memory, and Adam with learning rate memory.

\section{Experiments}
\label{sec:experiments}
We evaluate our meta learning algorithm in three different settings.
We start out with illustrating our approach on the Rosenbrock
function, a well known non-convex optimization problem. Then we show
extensive results on two learning tasks: First we investigate learning
binary classifiers on the MNIST data set, both illustrating the effect
of transferring the learning rate memory between similar tasks and
showcasing robustness to parameter choices. Finally, we extensively evaluate our meta-learning approach on online inverse dynamics learning tasks for manipulators.
We start out by explaining our experimental setup and the baseline methods we compare too.
\begin{figure*}[t]
  \centering
\includegraphics[trim=100 10 30 10, clip, width=0.31\linewidth]{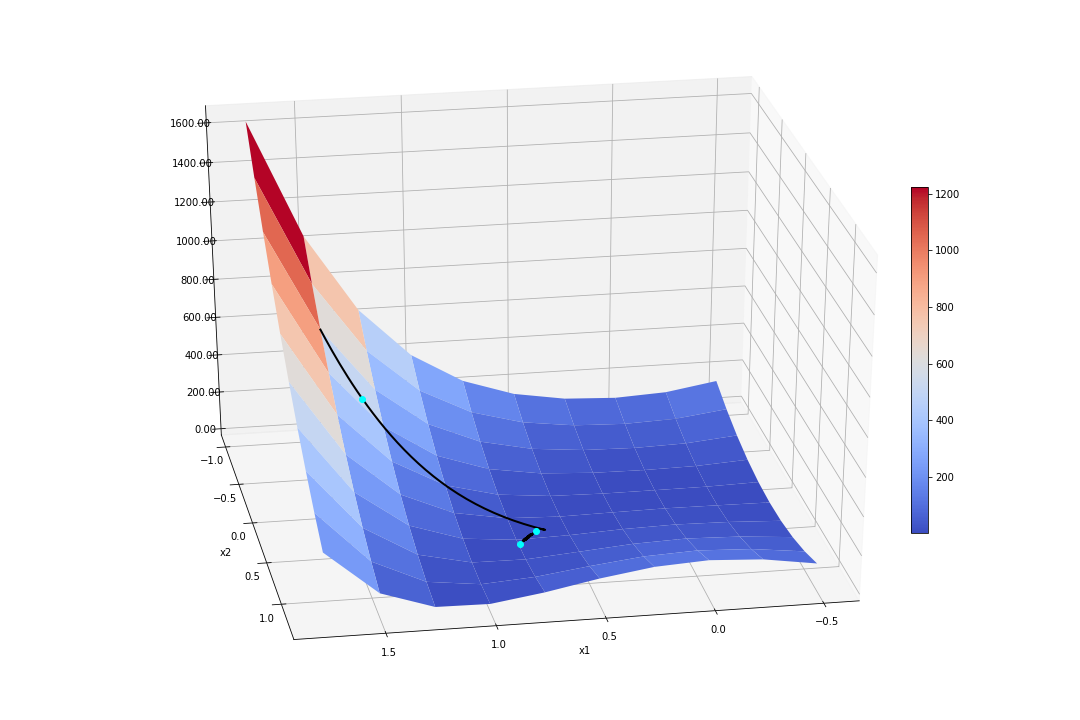}
\includegraphics[trim=20 10 30 10, clip, width=0.31\linewidth]{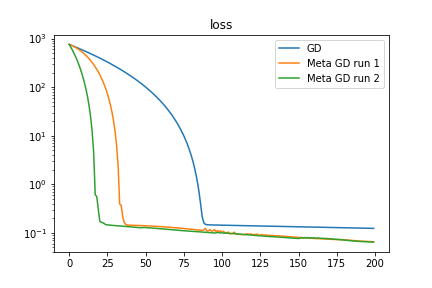}
\includegraphics[trim=0 10 40 10, clip, width=0.31\linewidth]{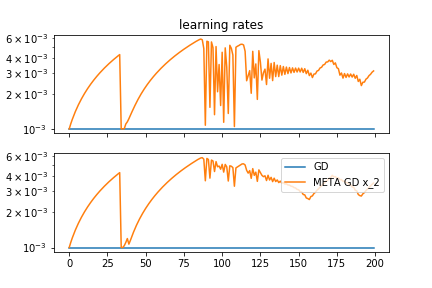}
\includegraphics[trim=20 10 30 20, clip, width=0.28\linewidth]{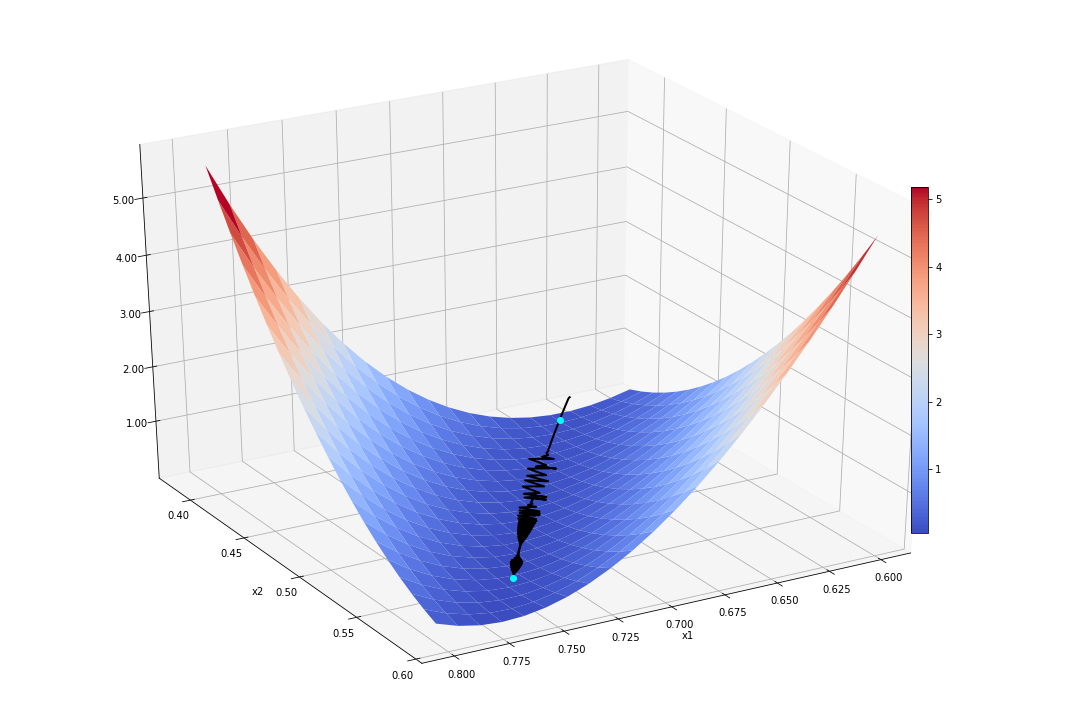}
\includegraphics[trim=10 10 60 20, clip,width=0.71\linewidth]{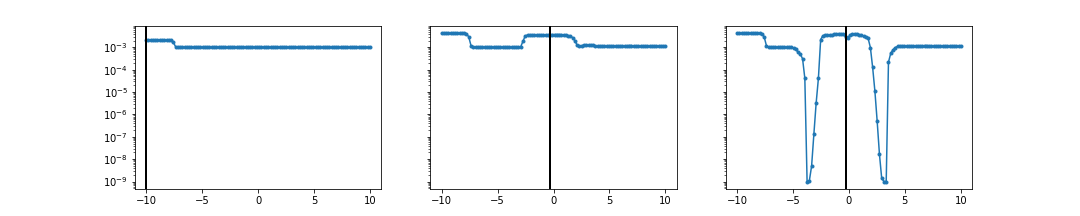}
\caption{\small {\bf top row}: (left) The Rosenbrock function with the
  meta-learners path. (middle) Convergence to minimum for gradient
  descent with $\eta=0.001$, and two consecutive runs of MetaGD with
  initial $\eta=0.001$. The first run with a learning rate memory,
  starts with all local models predicting the initial learning rate
  $\eta=0.001$, but quickly builds a model of learning rates - thus
  this already converges faster than without a memory. The second run
  of gradient descent with learning rate memory re-uses and updates
  the memory from run 1 and converges even faster. (right) the
  predicted learning rates for both dimensions, as a function of
  optimization steps. {\bf bottom row}: (left) zoomed in Rosenbrock
  function to visualize path in the valley. The blue dots (in both top
  and bottom Rosenbrock plots) indicate where we took a snapshot of
  the memories. (right) 3 snapshots of the memories, with x-axis
  representing values of the partial derivatives $z$, and the y-axis
  the predicted learning rate values. The black horizontal lines,
  indicate the current partial gradient - at the time of the
  snapshot.}
\label{fig:GD_vs_MetaGD_rb}
\end{figure*}
\subsection{Baselines and Experimental Setup}
As mentioned in Section~\ref{sec:main}, our proposed meta-learning can be combined with different base-optimizers. We need to choose an optimizer on both hierarchies, the task specific optimizer with base learning rate $\eta$, and the meta optimizer with base learning rate $\xi$. We present results for 3 variants: basic gradient descent without meta-learning (\gd), gradient descent with meta-learning with either gradient descent to update the memory (\metagd) or with Adam \cite{kingma2014adam} to update the memory (\metagdmemadam). For results involving neural networks, a memory of learning rates per hidden layer was learned.

We also compare to another meta-learner: \ltolbgd
\cite{andrychowicz2016learning} is a very recent approach which treats
the optimizer itself as a function approximator, typically represented
by a two layer recurrent LSTM network which transform the gradients
directly. \ltolbgd\ does require to learn the optimizer prior to
task-specific learning.
For our binary MNIST problem we sample initial neural network
configurations (without structural changes) and optimize on the same
data as used for evaluation. In order to avoid overfitting we run a
validation epoch with a random network instance after every second
optimization. After 16 rounds of optimizing a \ltolbgd\ learner we take
the best \ltolbgd\ network according to the validation evaluation.
All of the evaluated approaches, including our own, are based on
tensorflow \cite{abadi2016tensorflow} implementations.
\subsection{Rosenbrock Problem}
We start off by illustrating our approach on the Rosenbrock problem,
as visualized in Figure~\ref{fig:GD_vs_MetaGD_rb}(left). This is a 2D
optimization problem, and each dimension has it's own memory of learning rates (as shown
in Figure~\ref{fig:GD_vs_MetaGD_rb}). We compare the convergence of
gradient descent with $\eta=0.001$ and two consecutive runs of
gradient descent with a memory of learning rates, starting from the
same initial position. The memories are initialized to predict
$\eta=0.001$ for the first optimization run. For the second run, we carry
over the memories from the previous run.

The middle plot (top row) of Figure~\ref{fig:GD_vs_MetaGD_rb} shows
the convergence of each of these optimization runs. Notice how already
the first optimization run benefits from the online meta-learning. The second run converges even faster. The right plot (top row), shows the corresponding learning rates of both dimensions applied at each iteration of the second optimization run.

The bottom row shows a zoomed in Rosenbrock function, centered around
the path taken by the meta-learner. The blue dots indicate where we
have taken a snapshot of the learning rate memories. In the bottom
right - we show the memory of the second dimension, at those 3 different
time steps during the first run. We see how the memories initially
learn to predict increased learning rates\footnote{Note, this is the
  case because we clip the gradients at $\pm 10$ and thus it learns to
  increase the learning rates even in high curvature areas}. Once the
optimization hits the valley, it learns to predict larger learning
rates as long as the optimization stays within a certain gradient
range, at the borders of that gradient range the learning rate is
drastically reduced to prevent jumping out of the valley.

\subsection{Binary Mnist Problems}
\begin{figure*}
  \centering
\includegraphics[width=0.24\linewidth]{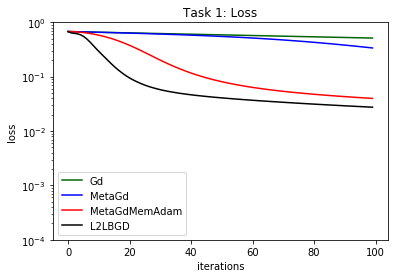}
\includegraphics[width=0.24\linewidth]{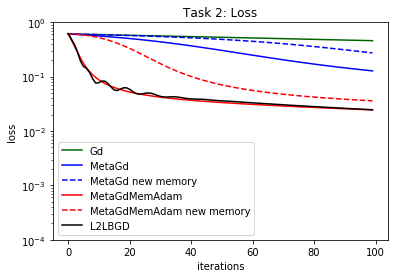}
\includegraphics[width=0.24\linewidth]{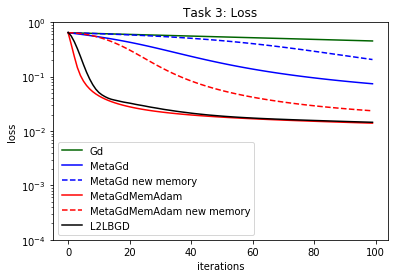}
\includegraphics[width=0.26\linewidth]{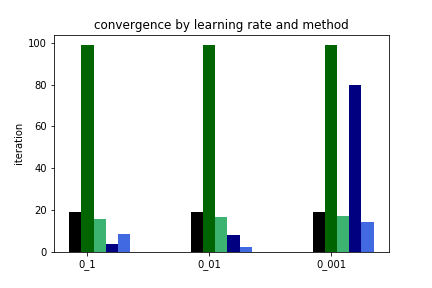}
\vspace{-0.2cm}
\caption{\small Results on the 3 binary MNIST learning tasks (left 3 plots top) convergence behavior of one randomly seeded (same for all optimizers) network on the 3 subsequent learning tasks with gradient descent based optimizers. (right): average number of iterations it took for the loss to drop below $0.1$ on task 1 for different initial learning rate choices ($[0.1, 0.01, 0.001]$). Results for meta-variants (from left to right: \ltolbgd, \gd, \metagd, \adam, \metagdmemadam) are obtained with new memories that have been initialized with the respective learning rate choice.
}
\label{fig:binary_mnist_memory_transfer}
\end{figure*}
In our second set of experiments we look at binary MNIST
classification tasks. Specifically we choose a sequence of 3 binary
classification learning tasks. This experimental setup allows us to
analyze the effect of transferring the memory of learning rates
between similar learning problems.
Task 1 learns to classify digits $1$ and $2$, task 2 digits $1$ and
$3$ and task 4 digit $1$ and $4$. Note, that we simply chose the first
$4$ digits for this experiment, and did not optimize that selection.

We use a neural network with the following structure: An input layer
that takes the image as input; a convolutional layer followed by a max
pooling operation; this is followed by a densely connected layer with
dropout; finally, the output layer is another dense layer. All hidden
units activation functions are rectified linear units. The loss
function is the softmax cross entropy loss. For all meta-learning
variants the memories (one per hidden layer) allocate a total of
$M=100$ local models. Training is performed in batch-mode.

We start out by illustrating the loss convergence of optimizing the
learning tasks in sequence in
Figure~\ref{fig:binary_mnist_memory_transfer} in the left three plots.
In green we see basic gradient descent without any memory of learning
rates. Convergence of \metagd\ and \metagdmemadam\ are shown in blue
and red respectively. For Task 1, no previous memory exists, thus all
optimizers start with the same learning rate (except \ltolbgd\ which
does not have an initial learning rate parameter). The meta variants
converge faster than basic gradient descent. In task 2 we deploy each
meta-variant with a new memory and with the memory trained in task 1.
Meta-variants with the new memory again convergence faster then basic
gradient descent; meta-variants initialized with the previously
learned memory converge even faster. Notice, that we continue to
update the memory of learning rates during task 2 optimization. For
task 3 we see the same convergence behavior as for task 2. We also
compare against \ltolbgd. This learned optimizer does not have a base
learning rate, it requires costly training on similar network
instantiations prior to usage. However, it requires a (meta) learning
rate to train the optimizer.
Here we chose to depict the best \ltolbgd\ optimizer we could train,
which was achieved with a learning rate of $0.01$.
Furthermore, the \ltolbgd\ optimizer was trained specifically for each task. 
Notice, on task 1 \ltolbgd\ achieves faster convergence, however any subsequent
learning tasks achieve similar convergence as the
\metagdmemadam-variant. 

Note - we have achieved similar results when using Adam as the base optimizer.
Adam itself, without any meta-learner, performs very well on this task, 
and can outperform all of the above meta-learning variants (including \ltolbgd). 
However, when combining Adam with our meta-learner, we achieve even better results. 
This confirms that our online meta-learning approach is versatile and can speed 
up convergence even when combined with an adaptive optimizer such as Adam.
These results are omitted due to space limitations.
Additionally, Adam does not perform well in the sequential problems 
such as the inverse dynamics task discussed in Sec.~\ref{sec:inv_dyn_online}.

In the right most plot of
Figure~\ref{fig:binary_mnist_memory_transfer}, we depict how many
iterations each optimizer variant requires to achieve an error below
$0.1$, as a function of initial learning rate values. These results
are averaged across 3 random seeds. Note \ltolbgd\ is constant because
after having been trained no initial learning rate parameter is
required. Here we include Adam, with and without meta-learning. In
green we see gradient descent \gd\ and \metagdmemadam, while in blue
we see \adam\ and \metagdmemadam. On average, our meta-learning
variants achieve a low error faster, irrespective of the initial
learning rate, while gradient descent does not converge within the
first 100 iterations, and Adam performs well if the learning rate is
chosen high enough. This confirms that at each iteration, our
meta-learning approach utilizes the observed data more efficiently and
as a result leads to faster learning progress. Furthermore, our
initial analysis also shows that with the use of our meta-learner the
learning frameworks convergence speed is less dependent on the choice
of the base learning rate. In the future, we hope to transfer this to
reinforcement learning settings.

\subsection{Inverse dynamics learning}
In our final set of experiments we explore the use of our
meta-learning variants on a typical motor control learning problem:
learning of an inverse dynamics model. Learning inverse dynamics is a
function approximation problem that maps the current joint position,
velocities and accelerations ($q, \dot{q}, \ddot{q}$) to torques
($\tau$).
This is an interesting learning problem since this function mapping
generally cannot be assumed to be stationary due to e.g. unknown
payload changes. Hence it requires online adaptation of the learned
inverse dynamics model, which can benefit from a meta-learner that is
invariant to such changes.
Another interesting aspect is the abundance of data, typically
continuously generated at very high rates (e.g. 1 KHz). Here we
consider two possibilities of processing such a high frequency data
stream: we either collect task-specific data and update a
task-specific inverse dynamics model in a batch setting
(Sec.~\ref{sec:inv_dyn_batch}); or we optimize the model online,
directly on the streaming data (Sec.~\ref{sec:inv_dyn_online}).

In both experiments we use a fixed-base manipulation platform equipped
with 7-DoF Kuka LWR IV arms and three-fingered Barrett Hands. We learn
the inverse dynamics model for the right arm resulting in a
21-dimensional input space with one output per joint.

\subsubsection{Batch Learning}\label{sec:inv_dyn_batch}
For this experiment we collect data of two motion tasks (in
simulation): Task 1 corresponds to acceleration policies
\cite{ijspeert2013dynamical} trained to move along a rectangular path
in the horizontal plane, while task 2 corresponds to the same
rectangular path in the vertical plane. These movements are performed
under strong perturbations of the assumed inverse dynamics model. Data
collection consists of recording joint position, velocities,
accelerations and torques at each time step (1ms). Each task is
trained on $10000$ collected data points. Per task, we train a neural
network per joint in a single batch setting, meaning that the full
dataset is used for each optimization step. The neural network
structure has 3 densely connected hidden layers with $[100, 50, 10]$
hidden units, respectively. All activation functions are rectified
linear units, and dropout prior to the output layer. The loss function
is the mean squared error (MSE) on the predicted torque values.

\begin{figure}
  \centering
\includegraphics[width=0.49\linewidth]{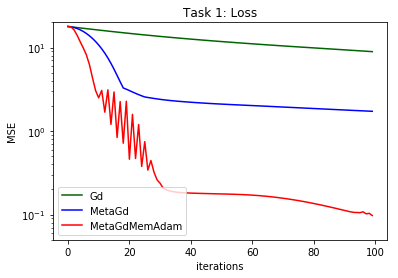}
\includegraphics[width=0.49\linewidth]{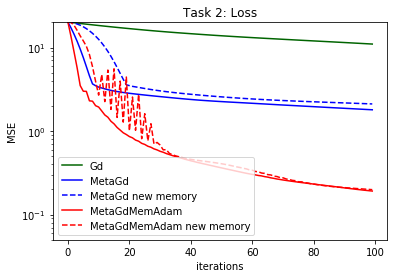}
\vspace{-0.1cm}
\caption{{\small Learning task-specific inverse dynamics models in
    batch mode, on the 1st joint of our 7-DOF manipulator.
    Meta-Learning helps to convergence faster within task 1 already,
    when transferring the learned memory of learning rates to task
    optimization, convergence speed is increased.}}
\label{fig:inv_dyn_results}
\end{figure}

We perform the two learning tasks in sequence: first we train a
dynamics model on task 1, then we learn a new (uninitialized) dynamics
model for task 2. The convergence of each optimization task is shown
in Figure~\ref{fig:inv_dyn_results}. Notice how using and optimizing
our meta-learner while optimizing the inverse dynamics model for task
1 already leads to faster convergence. When transferring the
meta-learner for task 2 learning convergence to a low error is even
faster. Thus, improving the convergence by re-using the meta-optimizer
while adapting to new tasks allows to perform faster incremental
learning of task specific inverse dynamics models.

\subsubsection{Sequential Online Learning}\label{sec:inv_dyn_online}
\begin{figure*}
  \centering
\includegraphics[trim=40 0 40 0,clip, width=0.325\linewidth]{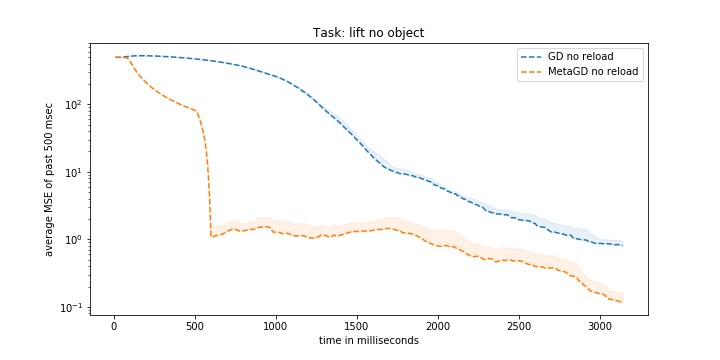}
\includegraphics[trim=40 0 40 0,clip, width=0.325\linewidth]{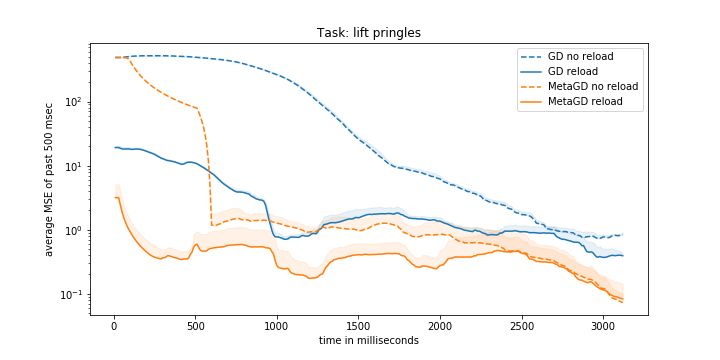}
\includegraphics[trim=40 0 50 0,clip, width=0.325\linewidth]{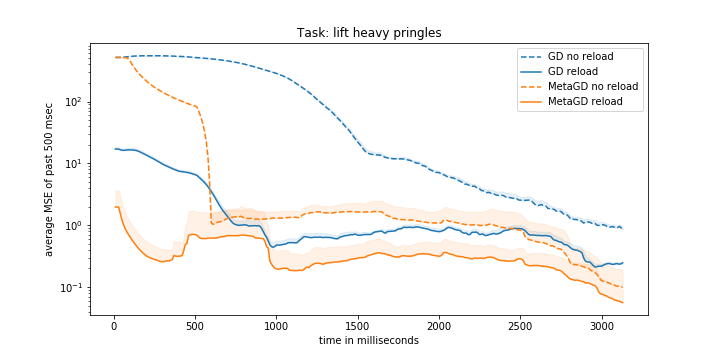}
\caption{\small Loss convergence when online learning a inverse
  dynamics model on the 1st joint of our 7-DOF manipulator. (left)
  uninitialized network is updated online with \gd and \metagd with an
  uninitialized memory. (middle) online learning convergence on
  task-variant 2 with and without reloading the network and
  meta-learner from task-variant 1. (right) convergence on
  task-variant 3 with and without reloading.}
\label{fig:inv_dyn_results_seq}
\end{figure*}
In this experiment we consider the task of lifting an object with
different configurations (no, light, heavy object) on the real
manipulation platform. We collect position, velocities, accelerations
and torques at each time step (1ms), resulting in three data sets with
more then 3000 samples each. Further, we execute each task variation
10 times to assess the variance of the learning process. All results
presented are averaged across these 10 trials.

In this experiment we have 3 learning phases, first we train an
uninitialized network and meta-learner on the \emph{lifting no object
  task}. Then this network is further trained on the data
corresponding to task-variant 2: \emph{light object}. Finally, the
same network is re-used and further adapted on task-variant 3:
\emph{heavy lifting}. This scenario tests, how quickly we can adapt
previously trained networks, with and without the meta-learner.

Again, we train one network per joint. The neural network structure
consists of 3 densely connected hidden layers with $[100, 50, 10]$
hidden units, respectively. All activation functions are rectified
linear units. The loss function used to optimize the parameters is the
mean-squared-error (MSE). For all experiments we use $M = 200$ local
models for each memory (per layer), a memory learning rate of $0.005$,
and gradient clipping of $1.0$. These values have been determined
empirically. We present results for joint 1 since it exhibits large
torque variations given payload changes. Notice, the initial high loss
values are due to the torque range which is between 30 and 35 N/m
hence the model optimized with mean squared error loss has to first
adjust for this offset. Different to the previous experiment each
learning phase is updating the network online using sequential data
batches of size 10, meaning we take data samples collected within the
last 10 milliseconds and perform one optimization step. Every data
sample is processed exactly once. This is a very challenging setting
since gradients are highly correlated and the overall optimization
step should be faster than 10 ms. From our empirical evaluation we
found that \adam\ does not perform well in these very correlated
settings which is why we omit the results. Computation times for both
\metagd\ and basic \gd\ are less then 3 ms on standard hardware,
hence, fast enough to consume a continuous data stream of inverse
dynamics data.

In Figure~\ref{fig:inv_dyn_results_seq} we illustrate the convergence
of the loss as we are moving along the trajectory of each lifting
task, when using $\eta=0.0001$. In the left most plot, we see
convergence on task 1 of \gd\ and \metagd. The middle plot shows
convergence on task 2 (light object), comparing learning from scratch
(no reloading of network or meta-learner) with continual learning of
the previously learned network (reload), for both basic \gd\ and
\metagd. As we can see, when warm-starting, we immediately start out with
a lower error for both \gd\ and \metagd. With \metagd\ we reduce the
experienced error faster. Note, even without reloading, \gd\ and
\metagd\ (net and memory uninitialized) manage to learn a good model
by the end of the movement, however they produce larger prediction
errors at the beginning of the movement. Similar behavior can be
observed on the final task of lifting the heavy object.

We performed this experiment for $\eta=[0.01, 0.001, 0.0001]$ and
summarize the mean loss for the first 500 ms of the task execution of
joint 1 in Table~\ref{tab:lift_1_3}. We are mostly interested in
assessing fast convergence, therefore the focus on the beginning of
the task. The data clearly illustrates the benefit of using our
proposed \metagd, that reaches a low loss faster, meaning using less data, in
almost all settings. Interesting to note is that with a really high
learning rate all methods achieve very good results e.g. $0.01$ heavy,
yet, due to the high learning rate there is a high chance that some
gradient in the sequence will drive the model into a ``bad'' parameter
space, resulting in poor convergence ($0.01$ no object).

%The loss evolution for the whole task sequence (learning rate 0.0001)
%is shown in Fig.~\ref{fig:inv_dyn_results_seq}. Here we can see the
%positive effect of both reusing the model and the \metagd\ which is
%able to adapt to new payload changes the fastest.
%
\begin{table}[]
    \centering
    \begin{footnotesize}
    \begin{tabular}{l|c|c|c|c|c}
 $\eta$ & & \gd & \metagd & \gd & \metagd \\
 &lift & noreload & noreload & reload &  reload \\
    \hline
    &no object                 & \textbf{180.062} & 228.421 & \textbf{180.062} & 228.421 \\
$0.01$ & light                  & 234.276 & 228.704 & 55.407 & \textbf{18.926} \\
    &  heavy            & 220.249 & 167.623 & 0.034 & \textbf{0.027} \\
\hline
    &no object                 & 119.317 & \textbf{65.221} & 119.317 & \textbf{65.221} \\
$0.001$ & light                  & 110.396 & 63.952 & 2.233 & \textbf{0.692} \\
    & heavy            & 113.62 & 66.958 & 0.618 & \textbf{0.454} \\
 \hline
    & no object                 & 468.849 & \textbf{80.219} & 468.849 & \textbf{80.219} \\
$0.0001$ & light                   & 489.441 & 78.799 & 10.712 & \textbf{0.593} \\
    & heavy            & 513.898 & 82.804 & 6.444 & \textbf{0.694} \\
\hline \hline
    &no object                 & 256.076& \textbf{124.62} & 256.076 & \textbf{124.62} \\
 average  & light                  & 278.038& 123.818 & 22.784 & \textbf{6.737} \\
    &  heavy            & 282.589& 105.795 & 2.365 & \textbf{0.392}
    \end{tabular}
    \caption{MSE of the first $500$ ms of each task execution. \vspace{-0.1cm}}
    \label{tab:lift_1_3}
    \end{footnotesize}
\end{table}

\section{Discussion and Future Work}
\label{sec:conclusion}
We have presented a novel meta-learning algorithm, that can
incrementally learn a memory of learning rates as a function of
current gradient observations. We have discussed how to learn this
memory in a computationally efficient manner and have shown that
deploying such a meta-learner leads to consistently faster convergence
in the number of iterations. Furthermore, we have shown that memories
of learning rates can be transferred between similar learning tasks,
and speed up convergence of the new -- previously unseen -- learning
problems.

However, thus far this effect has been constrained to base optimizers
that do not transform the gradient before updating the learning rate
memory. Thus, in future work we aim to investigate whether we can
extend the state with which the memory is indexed beyond simple
gradient information. The hope would be that this would allow the
memory to capture even more complex learning rate landscapes. It
further could enable to maintain the underlying structure between
tasks, better coping with forgetting of learning rate memories. The
challenge here is to do so while maintaining computational efficiency.

Another interesting avenue would be to use meta-learning for transfer learning.
This is especially interesting in robotics since 
real robot experiments are very costly and in general do
not scale in comparison to simulation. Yet, simulation results 
typically do not translate directly to the real world.
Thus, the meta-learner will hopefully result in less 
real robot experiments required to achieve good performance
since it can optimize the problem faster, thus, transferring information
from simulation to real world experiments.
Finally, we have shown how
a simple update rule as discussed in Section~\ref{sec:main} can create
a very effective meta-learner. Yet, it would be interesting to combine
our incremental meta-learner with even more expressive objectives that
can guide the learning of the memory.

%===============================================================================

% The maximum paper length is 8 pages excluding references and
% acknowledgements, and 9 pages including references and
% acknowledgements

% The acknowledgments are automatically included only in the final
% version of the paper.

%===============================================================================

\bibliographystyle{IEEEtran}
{\small
\bibliography{meta_learning}  % .bib
}
%\bibliographystyle{plain}
%\bibliographystyle{IEEEtran}
%\bibliography{../2017-survey-meta-learning/meta_learning}

\end{document}